\documentclass{article}
\usepackage[utf8]{inputenc}
\usepackage{amsmath}
\usepackage{amssymb}
\usepackage{nicefrac}
\usepackage{graphicx}
\usepackage{tikz}
\usetikzlibrary{shapes,decorations,arrows,calc,arrows.meta,fit,positioning}
\usepackage{caption}
\usepackage{subcaption}
\usepackage{xcolor}
\usepackage{multirow}
\usepackage{hyperref}
\usepackage[style=apa]{biblatex}
\newcommand{\CATE}{\text{CATE}}

\title{Personalized Decision Making - A Conceptual Introduction}
\author{Scott Mueller and Judea Pearl}
\date{April 3, 2022}

\begin{document}

\maketitle

\begin{abstract}
Personalized decision making targets the behavior of a specific individual, while population-based decision making concerns a sub-population resembling that individual. This paper clarifies the distinction between the two and explains why the former leads to more informed decisions. We further show that by combining experimental and observational studies we can obtain valuable information about individual behavior and, consequently, improve decisions over those obtained from experimental studies alone.
\end{abstract}

\section{Introduction}
The purpose of this paper is to provide a conceptual understanding of the distinction between personalized and population-based decision making, and to demonstrate both the advantages of the former and how it could be achieved.

Formally, this distinction is captured in the following two causal effects. Personalized decision making optimizes the Individual Treatment Effect (ITE):
\begin{equation}
    \text{ITE}(u) = Y(1,u)-Y(0,u)\label{eq:ite}
\end{equation}
where $Y(x,u)$ stands for the outcome that individual $u$ would attain had decision $x \in \{1, 0\}$ been taken. In contrast, population-based decision making optimizes the Conditional Average Treatment Effect (CATE):
\begin{equation}
    \text{CATE}(u) = E[Y(1,u') - Y(0,u') | C(u') = C(u)]\label{eq:cace}
\end{equation}
where $C(u)$ stands for a vector of characteristics observed on individual $u$, and the average is taken over all units $u'$ that share these characteristics.

We will show in this paper that the two objective functions lead to different decision strategies and that, although $\text{ITE}(u)$ is in general not identifiable, informative bounds can nevertheless be obtained by combining experimental and observational studies. We will further demonstrate how these bounds can improve decisions that would otherwise be taken using $\text{CATE}(u)$ as an objective function.

The paper is organized as follows. Section \ref{sec:prelim_example} will demonstrate, using an extreme example, two rather surprising findings. First, that population data are capable of providing decisive information on individual response and, second, that non-experimental data, usually discarded as bias-prone, can add information (regarding individual response) beyond that provided by a Randomized Controlled Trial (RCT) alone. Section \ref{sec:mot_example} will generalize these findings using a more realistic example, and will further demonstrate how critical decisions can be made using the information obtained and their ramifications to both the targeted individual and to a population-minded policy maker. Section \ref{sec:how_results} casts the findings of Section \ref{sec:mot_example} in a numerical setting, allowing for a quantitative appreciation of the magnitudes involved. This analysis leads to actionable policies that guarantee risk-free benefits in certain populations. Section \ref{sec:how_results} explains the mathematics that gives rise the results featured in Section \ref{sec:mot_example}, as well as the assumptions upon which they depend. Confounding in observational studies, typically problematic and in need of adjusting for, is shown to be helpful in narrowing the probabilities of benefit and harm. Section \ref{sec:monotonicity} discusses related topics such as monotonicity, number needed to treat and probability of harm. Finally, Section \ref{sec:related_works} provides annotated bibliography of the source papers, including related works.

For conceptual clarity, well-designed RCTs and observational studies are assumed throughout. As such we consider RCTs as having 100\% compliance and no selection bias or any other imperfections that often plague them (e.g., placebo effects). Similarly, observational studies are assumed to provide unbiased estimates of the statistical associations or conditional expectations they are designed to assess.

Trialists are usually suspicious of observational studies because the latter are either bias-prone or rely on subjective assumptions of ``no confounding'', which are hardly testable. Such trepidations do not apply to our analysis for two reasons. First, our analysis makes no modeling assumptions whatsoever when interpreting observational studies and, second, we actually benefit from the presence of confounding in the observational studies.

\section{Preliminary Semi-qualitative Example}\label{sec:prelim_example}
Our target of analysis is an individual response to a given treatment, namely, how an individual would react if given treatment and if denied treatment. Since no individual can be subjected to both treatment and its denial, its response function must be inferred from population data, originating from one or several studies. We are asking therefore: to what degree can population data inform us about an individual response?

Before tackling this general question, we wish to address two conceptual hurdles. First, why should population data provide \emph{any} information whatsoever on the individual response and, second, why should non-experimental data add any information (regarding individual response) to what we can learn with an RCT alone. The next simple example will demonstrate both points.

We conduct an RCT and find no difference between treatment (drug) and control (placebo), say 10\% in both treatment and control groups die, while the rest (90\%) survive. This makes us conclude that the drug is ineffective, but also leaves us uncertain between (at least) two competing models:
\begin{itemize}
    \item Model-1 -- The drug has no effect whatsoever on any individual and
    \item Model-2 -- The drug saves 10\% of the population and kills another 10\%.
\end{itemize}

From a policy maker viewpoint the two models may be deemed equivalent, the drug has zero average effect on the target population. But from an individual viewpoint the two models differ substantially in the sets of risks and opportunities they offer. According to Model-1, the drug is useless but safe. According to Model-2, however, the drug may be deemed dangerous by some and a life-saver by others.

To see how such attitudes may emerge, assume, for the sake of argument, that the drug also provides temporary pain relief. Model-1 would be deemed desirable and safe by all, whereas Model-2 will scare away those who do not urgently need the pain relief, while offering a glimpse of hope to those whose suffering has become unbearable, and who would be ready to risk death for the chance (10\%) of recovery. (Hoping, of course, they are among the lucky beneficiaries.)

This simple example will also allow us to illustrate the second theme of our paper – the crucial role of observational studies. We will now show that supplementing the RCT with an observational study on the same population (conducted, for example, by an independent survey of patients who have the option of taking or avoiding the drug) would allow us to decide between the two models, totally changing our understanding of what risks await an individual taking the drug.

Consider an extreme case where the observational study shows 100\% survival in both drug-choosing and drug-avoiding patients, as if each patient knew in advance where danger lies and managed to avoid it.  Such a finding, though extreme and unlikely, immediately rules out Model-1 which claims no treatment effect on any individual. This is because the mere fact that patients succeed 100\% of the time to avoid harm where harm does exist (revealed through the 10\% death in the randomized trial) means that choice makes a difference, contrary to Model-1's claim that choice makes no difference.

The reader will surely see that the same argument applies when the probability of survival among option-having individuals is not precisely 100\% but simply higher (or lower) than the probability of survival in the RCT. Using the RCT study alone, in contrast, we were unable to rule out Model-1, or even to distinguish Model-1 from Model-2.

We now present another edge case where Model-2, rather than Model-1, is ruled out as impossible. Assume the observational study informs us that all those who chose the drug died and all who avoided the drug survived. It seems that drug-choosers were truly dumb while drug-avoiders knew precisely what's good for them. This is perfectly feasible, but it also tells us that no one can be \emph{cured} by the drug, contrary to the assertion made by Model-2, that the drug cures 10\% and kills 10\%. To be cured, a person must survive if treated and die if not treated. But none of the drug-choosers could have been cured, because they all died, and none of the drug avoiders could have been cured because they all survived (they might have survived had they taken the drug, but then it would not have been the drug that cured them).  Thus, Model-2 cannot explain these observational results, and must be ruled out.

Now that we have demonstrated conceptually how certain combinations of observational and experimental data can provide information on individual behavior that each study alone cannot, we are ready to go to a more realistic motivating example which, based on theoretical bounds derived in \cite{tian2000}, establishes individual behavior for any combination of observational and experimental data\footnote{The example we will work out happened to be identifiable due to particular combinations of data, though, in general, the data may not permit point estimates of individual causal effects} and, moreover, demonstrates critical decision making ramifications of the information obtained.

\section{Motivating Numerical Example}\label{sec:mot_example}
The next example to be considered deals with the effect of a drug on two subpopulations, males and females. Unlike the extreme case considered in Section \ref{sec:prelim_example}, the drug is found to be somewhat effective for both males and females and, in addition, deaths are found to occur in the observational study as well. We will demonstrate that, although men and women are totally indistinguishable in the RCT study, adding observational data proves men to react markedly different than women, calling for two different treatment policies in the two groups. Whereas a woman has a 28\% chance of benefiting from the drug and no danger at all of being harmed by it, a man has a 49\% chance of benefiting from it and as much as a 21\% chance of dying because of it.

To cast the story in a realistic setting, we imagine the testing of a new drug, aimed to help patients suffering from a deadly disease. An RCT is conducted to evaluate the efficacy of the drug and is found to be 28\% effective\footnote{To simplify matters, we are treating each experimental study data as an ideal RCT, with 100\% compliance and no selection bias or any other biases that often plague RCTs.} in both males and females. In other words $\text{CATE}(\text{male}) = \text{CATE}(\text{female}) = 0.28$. The drug is approved and, after a year of use, a follow up randomized study is conducted yielding the same results; namely CATE remained 0.28, and men and women remained totally indistinguishable in their responses, as shown in Table \ref{tab:fem_male_cace}.

\begin{table}[h]
    \centering
    \begin{tabular}{l|r|r}
         & \multicolumn{1}{|c|}{Female Survivals} & \multicolumn{1}{|c}{Male Survivals}\\
         \hline
        $do(\text{drug})$ & 489/1000 (49\%) & 490/1000 (49\%)\\ 
        $do(\text{no drug})$ & 210/1000 (21\%) & 210/1000 (21\%)\\
        \hline
        CATE & 28\%\ \  & 28\%\ \ \\
    \end{tabular}
    \caption{Female vs male CATE}
    \label{tab:fem_male_cace}
\end{table}

Let us focus on the second RCT (Table \ref{tab:fem_male_cace}), since the first was used for drug approval only, and its findings are the same as the second. The RCT tells us that there was a $28\%$ improvement, on average, in taking the drug compared to not taking the drug. This was the case among both females and males: $\text{CATE}(\text{female}) = \text{CATE}(\text{male}) = 0.28$, where $do(\text{drug})$ and $do(\text{no-drug})$ are the treatment and control arms in the RCT. It thus appears reasonable to conclude that the drug has a net remedial effect on some patients and that every patient, be it male or female, should be advised to take the drug and benefit from its promise of increasing one's chances of recovery (by $28\%$).

At this point, the drug manufacturer ventured to find out to what degree people actually buy the approved drug, following its recommended usage. A market survey was conducted (observational study) and revealed\footnote{As with the experimental studies, observational studies are assumed to provide unbiased estimates for simplicity.} that only $70\%$ of men and $70\%$ of women actually chose to take the drug; problems with side effects and rumors of unexpected deaths may have caused the other $30\%$ to avoid it. A careful examination of the observational study has further revealed substantial differences in survival rates of men and women who chose to use the drug (shown in Tables \ref{tab:results_fem} and \ref{tab:results_male}). The rate of recovery among drug-choosing men was exactly the same as that among the drug-avoiding men ($70\%$ for each), but the rate of recovery among drug-choosing women was $43\%$ lower than among drug-avoiding women ($0.27$ vs $0.70$, in Table \ref{tab:results_fem}). It appears as though many women who chose the drug were already in an advanced stage of the disease, which may account for their low recovery rate of $27\%$.

\begin{table}[h]
    \centering
    \begin{tabular}{l|l|r|r|r}
        \multicolumn{2}{}{} & \multicolumn{1}{|c|}{Survivals} & \multicolumn{1}{|c|}{Deaths} & \multicolumn{1}{|c}{Total}\\
        \hline
        \multirow{2}{*}{Experimental} & $do(\text{drug})$ & 489 (49\%) & 511 (51\%) & 1,000 (50\%)\\
         & $do(\text{no drug})$ & 210 (21\%) & 790 (79\%) & 1,000 (50\%)\\
        \hline
        \multirow{2}{*}{Observational} & drug & 378 (27\%) & 1,022 (73\%) & 1,400 (70\%)\\
         & no drug & 420 (70\%) & 180 (30\%) & 600 (30\%)\\
    \end{tabular}
    \caption{Female survival and recovery data}
    \label{tab:results_fem}
\end{table}
\begin{table}[h]
    \centering
    \begin{tabular}{l|l|r|r|r}
        \multicolumn{2}{}{} & \multicolumn{1}{|c|}{Survivals} & \multicolumn{1}{|c|}{Deaths} & \multicolumn{1}{|c}{Total}\\
        \hline
        \multirow{2}{*}{Experimental} & $do(\text{drug})$ & 490 (49\%) & 510 (51\%) & 1,000 (50\%)\\
         & $do(\text{no drug})$ & 210 (21\%) & 790 (79\%) & 1,000 (50\%)\\
        \hline
        \multirow{2}{*}{Observational} & drug & 980 (70\%) & 420 (30\%) & 1,400 (70\%)\\
         & no drug & 420 (70\%) & 180 (30\%) & 600 (30\%)\\
    \end{tabular}
    \caption{Male survival and recovery data}
    \label{tab:results_male}
\end{table}

At this point, having data from both experimental and observational studies we can estimate the individual treatment effects for both a typical man and a typical woman. Quantitative analysis shows (see Section \ref{sec:how_results}) that, with the data above, the drug affects men markedly differently from the way it affects women. Whereas a woman has a 28\% chance of benefiting from the drug and no danger at all of being harmed by it, a man has a 49\% chance of benefiting from it and as much as a 21\% chance of dying because of it --- a serious cause for concern. Note that based on the experimental data alone (Table \ref{tab:fem_male_cace}), no difference at all can be noticed between men and women.

The ramifications of these findings on personal decision making are enormous. First, they tell us that the drug is not as safe as the RCT would have us believe, it may cause death in a sizable fraction of patients. Second, they tell us that a woman is totally clear of such dangers, and should have no hesitation to take the drug, unlike a man, who faces a decision; a 21\% chance of being harmed by the drug is cause for concern. Physicians, likewise, should be aware of the risks involved before recommending the drug to a man. Third, the data tell policy makers what the overall societal benefit would be if the drug is administered to women only; 28\% of the drug-takers would survive who would die otherwise. Finally, knowing the relative sizes of the benefiting vs harmed subpopulations swings open the door for finding the mechanisms responsible for the differences as well as identifying measurable markers that characterize those subpopulations.

For example:
\begin{itemize}
    \item Our analysis has identified ``Sex'' to be an important feature, separating those who are harmed from those saved by the drug. In the same way we can leverage other measured features, say family history, a genetic marker, or a side-effect, and check whether they shrink the sizes of the susceptible subpopulations. The results would be a set of features that approximate responses at the individual level. Note again that absent observational data and a calculus for combining them with the RCT data, we would not be able to identify such informative features. A feature like ``Sex'' would be deemed irrelevant, since men and women were indistinguishable in our RCT studies.
    \item Our ability to identify relevant informative features as described above can be leveraged to amplify the potential benefits of the drug. For example, if we identify a marker that characterizes men who would die only if they take the drug and prevent those patients from taking the drug, the drug would cure 62\% of male patients who would be allowed to use it. This is because we don't administer the drug to the 21\% who would've been killed by the drug. Those patients will now survive, so a total of 70\% of patients will be cured because of this combination of marker identification and drug administration. This unveils an enormous potential of the drug at hand, which was totally concealed by the 28\% effectiveness estimated in the RCT studies.
\end{itemize}

\section{How the Results Were Obtained}\label{sec:how_results}
For the purpose of analysis, let us denote $y_t$ as recovery among the RCT treatment group and $y_c$ as recovery among the RCT control group. The causal effects for treatment and control groups, $P(y_t|\text{Gender})$ and $P(y_c|\text{Gender})$, were the same\footnote{$P(y_t|\text{female})$ was rounded up from $48.9\%$ to $49\%$. The $0.001$ difference between $P(y_t|\text{female}$) and $P(y_t|\text{male})$ wasn't necessary, but was constructed to allow for clean point estimates.}, no differences were noted between males and females.

In addition to the above RCT, we posited an observational study (survey) conducted on the same population. Let us denote $P(y|t, \text{Gender})$ and $P(y|c, \text{Gender})$ as recovery among the drug-choosers and recovery among the drug-avoiders, respectively.

With this notation at hand, our problem is to compute the probability of benefit
\begin{equation}
      P(\text{benefit}) = P(y_t, y'_c)\label{eq:benefit}
\end{equation}
from the following data sources: $P(y_t)$, $P(y_c)$, $P(y|t)$, $P(y|c)$, and $P(t)$. The first two denote the data obtained from the RCT and the last three, data obtained from the survey. Non-recovery is represented by $y'$, so $y'_c$ is non-recovery among the RCT control group. Eq. \eqref{eq:benefit} should be interpreted as the probability that an individual would both recover if assigned to the RCT treatment arm and die if assigned to control\footnote{Tian and Pearl (\cite{tian2000}) called $P(\text{benefit})$ ``Probability of Necessity and Sufficiency'' (PNS). The relationship between PNS and ITE \eqref{eq:ite} is explicated in Section \ref{sec:related_works}}.

The results of the observational and experimental studies are not independent of each other since, barring selection bias, participants in the two studies are selected from the same overall population, ideally consisting of the eventual users of the drug. At the individual level, the connection between behaviors in the two studies relies on an assumption known as \emph{consistency} (\cite{pearl2009causality,pearl2010})\footnote{Consistency is a property imposed at the individual level, often written as
\begin{equation*}
    Y = X \cdot Y(1) + (1-X) \cdot Y(0)
\end{equation*}
for binary X and Y. Rubin (\cite{rubin1974}) considered consistency to be an assumption in SUTVA, which defines the potential outcome (PO) framework. Pearl (\cite{pearl2010}) considered consistency to be a theorem of Structural Equation Models, a violation of which reflects imperfections (e.g. placebo effects) in RCT practices.}, asserting that an individual response to treatment depends entirely on biological factors, unaffected by the settings in which treatment is taken\footnote{In medical practices, clinical experts rarely rely on the assumption of biological equivalence. The very participation in a study tends to create fears and expectations that affect patients response to treatment. Moreover, selection bias (\cite{bareinboim2014}) is a major problem in clinical trials, since subjects are recruited by stringent health criteria and, unlike those in observational studies, they must undergo consent procedures. For these two reasons, RCT practitioners compare only patients that undergo the same recruitment procedure and, accordingly, report only the difference $P(y_t)-P(y_c)$. More elaborate procedures (\cite{bareinboim2014}) must be deployed to overcome both selection bias and placebo effects when experimental and observational studies are to be combined.}. In other words, the outcome of a person choosing the drug would be the same had this person been assigned to the treatment group in an RCT study. Similarly, if we observe someone avoiding the drug, their outcome is the same as if they were in the control group of our RCT.



In terms of our notation, consistency implies:
\begin{equation}
    P(y_t|t)= P(y|t), P(y_c|c)= P(y|c).\label{eq:consistency}
\end{equation}
In words, the probability that a drug-chooser would recover in the treatment arm of the RCT, $P(y_t|t)$, is the same as the probability of recovery in the observational study, $P(y|t)$.

Based on this assumption, and leveraging both experimental and observational data, Tian and Pearl (\cite{tian2000}) derived the following tight bounds on the probability of benefit, as defined in \eqref{eq:benefit}:
\begin{equation}
    \max\left\{\begin{array}{c}
        0,\\
        P(y_t) - P(y_c),\\
        P(y) - P(y_c),\\
        P(y_t) - P(y)\\
    \end{array}\right\} \leqslant P(\text{benefit}) \leqslant \min\left\{\begin{array}{c}
        P(y_t),\\
        P(y'_c),\\
        P(t,y) + P(c,y'),\\
        P(y_t) - P(y_c)\ +\\
        \ P(t, y') + P(c, y)
    \end{array}\right\}.\label{eq:tian}
\end{equation}
Here $P(y'_c)$ stands for $1-P(y_c)$, namely the probability of death in the control group. The same bounds hold for any subpopulation, say males or females, if every term in \eqref{eq:tian} is conditioned on the appropriate class.

Applying these expressions to the female data from Table \ref{tab:results_fem} gives the following bounds on $P(\text{benefit}|\text{female})$:
\begin{align}
    \max\{0, 0.279, 0.09, 0.189\} &\leqslant P(\text{benefit}|\text{female}) \leqslant \min\{0.489,0.79,0.279,1\},\nonumber\\
    0.279 &\leqslant P(\text{benefit}|\text{female}) \leqslant 0.279.\label{eq:benefit_female}
\end{align}

Similarly, for men we get:
\begin{align}
    \max\{0, 0.28, 0.49, -0.21\} &\leqslant P(\text{benefit}|\text{male}) \leqslant \min\{0.49, 0.79, 0.58, 0.7\},\nonumber\\
    0.49 &\leqslant P(\text{benefit}|\text{male}) \leqslant 0.49.\label{eq:benefit_male}
\end{align}

Thus, the bounds for both females and males, in \eqref{eq:benefit_female} and \eqref{eq:benefit_male}, collapse to point estimates:
\begin{align*}
    P(\text{benefit}|\text{female}) &= 0.279,\\
    P(\text{benefit}|\text{male}) &= 0.49.
\end{align*}

We aren't always so fortunate to have a complete set of observational and experimental data at our disposal. When some data is absent, we are allowed to discard arguments to $\max$ or $\min$ in \eqref{eq:tian} that depend on that data. For example, if we lack all experimental data, the only applicable lower bound in \eqref{eq:tian} is $0$ and the only applicable upper bound is $P(t, y) + P(c, y')$:
\begin{equation}
    0 \leqslant P(\text{benefit}) \leqslant P(t, y) + P(c, y').\label{eq:benefit_bounds_obs}
\end{equation}
Applying these observational data only bounds to males and females yields:
\begin{align*}
    0 &\leqslant P(\text{benefit}|\text{female}) \leqslant 0.279,\\
    0 &\leqslant P(\text{benefit}|\text{male}) \leqslant 0.58.
\end{align*}
Naturally, these are far more loose than the point estimates when combined experimental and observational data is fully available. Let's similarly examine what can be computed with purely experimental data. Without observational data, only the first two arguments to $\max$ of the lower bound and $\min$ of the upper bound of $P(\text{benefit})$ in \eqref{eq:tian} are applicable:
\begin{equation}
    \max\{0,P(y_t)-P(y_c)\} \leqslant P(\text{benefit}) \leqslant \min\{P(y_t),P(y'_c)\}.\label{eq:benefit_bounds_exp}
\end{equation}
Applying these experimental data only bounds to males and females yields:
\begin{align*}
    0.279 &\leqslant P(\text{benefit}|\text{female}) \leqslant 0.489,\\
    0.28 &\leqslant P(\text{benefit}|\text{male}) \leqslant 0.49.
\end{align*}
Again, these are fairly loose bounds, especially when compared to the point estimates obtained with combined data. Notice that the overlap between the female bounds using observational data, $0 \leqslant P(\text{benefit}|\text{female}) \leqslant 0.279$, and the female bounds using experimental data, $0.279 \leqslant P(\text{benefit}|\text{female}) \leqslant 0.489$ is the point estimate $P(\text{benefit}|\text{female}) = 0.279$. The more comprehensive Tian-Pearl bounds formula \eqref{eq:tian} wasn't necessary. However, the intersection of the male bounds using observational data, $0 \leqslant P(\text{benefit}|\text{male}) \leqslant 0.58$, and the male bounds using experimental data, $0.28 \leqslant P(\text{benefit}|\text{male}) \leqslant 0.49$, does not provide us with narrower bounds. For males, the comprehensive Tian-Pearl bounds in \eqref{eq:tian} was necessary for narrow bounds (in this case, a point estimate).

Having seen this mechanism of combining observational and experimental data in \eqref{eq:tian} work so well, the reader may ask what's behind this? The intuition comes from the fact that observational data incorporates individuals' whims, and whims are proxies for hidden factors that may affect that individual's response to treatments. Such ``confounding'' factors are usually problematic in causal inference, since they lead to biased conclusions, sometimes completely reversing a treatment's effect (\cite{pearl2014simpson}). Confounding then needs to be adjusted for. However, here confounding helps us, exposing the underlying mechanisms its associated whims and desires are a proxy for.


Finally, as noted in Section \ref{sec:mot_example}, knowing the relative sizes of the benefiting vs harmed subpopulations demands investment in finding mechanisms responsible for the differences as well as characterizations of those subpopulations. For example, women above a certain age may be affected differently by the drug, to be detected by how age affects the bounds on the individual response. Such characteristics can potentially be narrowed repeatedly until the drug's efficacy can be predicted for an individual with certainty or the underlying mechanisms of the drug can be fully understood.

None of this was possible with only the RCT. Yet, remarkably, an observational study, however sloppy and uncontrolled, provides a deeper perspective on a treatment's effectiveness. It incorporates individuals' whims and desires that govern behavior under free-choice settings. And, since such whims and desires are often proxies for factors that also affect outcomes and treatments (i.e., confounders), we gain additional insight hidden by RCTs.

\section{Monotonicity, Probability of Harm, Number needed to Treat, and Other Results}\label{sec:monotonicity}
A natural question to ask at this point is, under what condition will RCT results constitute a point estimate for our target quantity, $P(\text{benefit})$? Pearl (\cite{pearl1999probabilities}) has shown that this occurs under a condition called \emph{monotonicity}, namely, when the treatment cannot harm any individual, formally
\begin{equation*}
    P(y_t, y'_c) = P(\text{harm}) = 0.
\end{equation*}
This can be shown through a general relationship between $P(\text{harm})$, $P(\text{benefit})$, and ATE, which reads\footnote{Eq. \eqref{eq:harm_from_ace} can be obtained by expanding ATE, subtracting $P(y_c) = P(y_c, y_t) + P(y_c, y'_t)$ from $P(y_t) = P(y_t,y_c) + P(y_t, y'_c)$ to obtain $\text{ATE} = P(y_t, y'c) - P(y'_t, y_c) = P(\text{benefit}) - P(\text{harm})$.}:
\begin{equation}
    P(\text{harm}) = P(\text{benefit}) - \text{ATE}.\label{eq:harm_from_ace}
\end{equation}

Eq. \eqref{eq:harm_from_ace} can serve two purposes. First, it tells us immediately that under monotonicity (i.e., $P(\text{harm}) = 0$), $P(\text{benefit})$ coincides with ATE, or, in other words, ATE constitutes a point estimate of $P(\text{benefit})$. Second, it allows us to compute $P(\text{harm})$ from $P(\text{benefit})$ and ATE in cases where monotonicity does not hold, as was the case for men in the numeric example of Section \ref{sec:mot_example}.

For each of females and males, in the above example, their respective $P(\text{benefit})$ and ATE are known. Therefore their probabilities of harm are known as well:
\begin{align*}
    P(\text{harm}|\text{female}) &= P(\text{benefit}|\text{female}) - \CATE(\text{female})\\
    &= 0.279 - 0.279 = 0,\\
    P(\text{harm}|\text{male}) &= P(\text{benefit}|\text{male}) - \CATE(\text{male})\\
    &= 0.49 - 0.28 = 0.21.
\end{align*}

Another concept that has become popular among trialists is ``Number Needed to Treat'' (NNT)\footnote{NNT isn't without controversy. Issues revolve around cases where confidence intervals for ATE include $0$ rendering NNT undefined. If the ATE is in fact $0$ then the explanatory benefit of NNT can easily get lost. Stovitz and Shrier show why baseline risk is important for medical decision making if NNT is relied upon (\cite{stovitz2013}).} which is defined as: ``The number of persons needed to be treated, on average, to prevent one more event (e.g., occurrence of a disease to be prevented, complication, adverse reaction, relapse)'' (\cite{porta2016}). Indeed, the phrase ``number of persons needed to be treated'' translates the academic notion of treatment efficacy into a vivid scenario that is clinically a more meaningful way of expressing the benefit of one intervention over another. Unfortunately, generations of trialists have failed to notice the counterfactual nature of the verb ``prevent'', and have estimated NNT as the inverse of ATE (\cite{vancak2020}):
\begin{equation}
    \text{NNT} = \frac{1}{P(y_t) - P(y_c)}\label{eq:nnt_classic}
\end{equation}
instead of the inverse of $P(\text{benefit})$:
\begin{equation}
    \text{NNT} = \frac{1}{P(\text{benefit})}.
\end{equation}
Eq. \eqref{eq:nnt_classic} has been used indiscriminately including cases where treatment may cause harm to some individuals. In such cases, NNT should be estimated as bounds, specifically the inverse of Equations \eqref{eq:tian}, \eqref{eq:benefit_bounds_obs}, and \eqref{eq:benefit_bounds_exp}. For example, if only experimental data are available, Equation \eqref{eq:nnt_classic} merely provides an upper bound:
\begin{equation}
    \max\left\{\frac{1}{P(y_t)}, \frac{1}{P(y'_c)}\right\} \leqslant \text{NNT} \leqslant \frac{1}{P(y_t) - P(y_c)}
\end{equation}
and the lower bound is provided by Eq. \eqref{eq:benefit_bounds_exp}.

Given its ubiquity in interpreting experimental studies, a natural question to ask is whether monotonicity is testable. This question can be answered by examining the bounds on $P(\text{harm})$ and asking what conditions would guarantee an upper bound of $0$. The bounds on the probability of harm are:
\begin{equation}
    \max\left\{\begin{array}{c}
        0,\\
        P(y_c) - P(y_t),\\
        P(y) - P(y_t),\\
        P(y_c) - P(y)\\
    \end{array}\right\} \leqslant P(\text{harm}) \leqslant \min\left\{\begin{array}{c}
        P(y_c),\\
        P(y'_t),\\
        P(t,y') + P(c,y),\\
        P(y_c) - P(y_t)\ +\\
        \ \ \ \ P(t, y) + P(c, y')
    \end{array}\right\}.\label{eq:harm_bounds}
\end{equation}

We see that, when $P(y_t) > P(y_c)$, the sufficient test demands that any of the following pathological conditions be true:
\begin{align}
    P(y_c) &= 0\text{ or}\label{eq:suff_test_mono_yc}\\
    P(y_t) &= 1\text{ or}\label{eq:suff_test_mono_yt}\\
    P(t,y') = P(c,y) &= 0.\label{eq:suff_test_mono_obs}
\end{align}

The necessary test for monotonicity is more informative and is given in Causality (\cite[p. 294]{pearl2009causality}):
\begin{equation}
    P(y_t) \geqslant P(y) \geqslant P(y_c).\label{eq:nec_test_mono}
\end{equation}
This test is useful for two reasons. First, it can quickly eliminate the possibility of monotonicity by checking for a violation of \eqref{eq:nec_test_mono}. Second, such a violation indicates a high variability among individuals in the subpopulation considered, which, in turn, calls for a search for the mechanism responsible for the variability.

\section{Annotated Bibliography for Related Works}\label{sec:related_works}
The following is a list of papers that analyze probabilities of causation and lead to the results reported above.
\begin{itemize}
    \item Chapter 9 of Causality (\cite{pearl2009causality}) derives bounds on individual-level probabilities of causation and discusses their ramifications in legal settings. It also demonstrates how the bounds collapse to point estimates under certain combinations of observational and experimental data.
    \item (\cite{tian2000}) develops bounds on individual level causation by combining data from experimental and observational studies. This includes Probability of Sufficiency (PS), Probability of Necessity (PN), and Probability of Necessity and Sufficiency (PNS). PNS is equivalent to $P(\text{benefit})$ above. $\text{PNS}(u) = P(\text{benefit}|u)$, the probability that individual $U=u$ survives if treated and does not survive if not treated, is related to $\text{ITE}(u)$ \eqref{eq:ite} via the equation:
\begin{equation}
    \text{PNS}(u) = P(\text{ITE}(u') > 0 | C(u') = C(u)).\label{eq:pns}
\end{equation}

In words, $\text{PNS}(u)$ equals the proportion of units $u'$ sharing the characteristics of $u$ that would positively benefit from the treatment. The reason is as follows. Recall that (for binary variables) $\text{ITE}(u)$ is $1$ when the individual benefits from the treatment, $\text{ITE}(u)$ is $0$ when the individual responds the same to either treatment, and $\text{ITE}(u)$ is $-1$ when the individual is harmed by treatment. Thus, for any given population, $\text{PNS} = P(\text{ITE}(u) > 0)$. Focusing on the sub-population of individuals $u'$ that share the characteristics of $u$, $C(u') = C(u)$, we obtain \eqref{eq:pns}. In words, $\text{PNS}(u)$ is the fraction of indistinguishable individuals that would benefit from treatment. Note that whereas \eqref{eq:cace} is can be estimated by controlled experiments over the population $C(u')=C(u)$, \eqref{eq:pns} is defined counterfactually, hence, it cannot be estimated solely by such experiments; it requires additional ingredients as described in the text below.
    \item (\cite{mueller2020}) provides an interactive visualization of individual level causation, allowing readers to observe the dynamics of the bounds as one changes the available data.
    \item (\cite{li2019}) optimizes societal benefit of selecting a unit $u$, when provided costs associated with the four different types of individuals, benefiting, harmed, always surviving, and doomed.
    \item (\cite{mueller2021}) takes into account the causal graph to obtain narrower bounds on PNS. The hypothetical study in this article was able to calculate point estimates of PNS, but often the best we can get are bounds.
    \item (\cite{pearl2015}) demonstrates how combining observational and experimental data can be informative for determining Causes of Effects, namely, assessing the probability PN that one event was a necessary cause of an observed outcome.
    \item (\cite{dawid2022effects}) analyze Causes of Effects (CoE), defined by PN, the probability that a given intervention is a necessary cause for an observed outcome.  Dawid and Musio further analyze whether bounds on PN can be narrowed with data on mediators.
\end{itemize}

\section{Conclusion}
One of the least disputed mantra of causal inference is that we cannot access individual causal effects; we can observe an individual response to treatment or to no-treatment but never to both. However, our theoretical results show that we can get bounds on individual causal effects, which sometimes can be quite narrow and allow us to make accurate personalized decisions. Conditioning on additional characteristics of the individual involved should provide, of course, additional person-specific information. However, such additions are accompanied with increased variance and must therefore be limited by the sample size available in each stratum. Our bounds are not subject to this limitation and takes full advantage of the large sample size usually available in observational studies. We project therefore that these methods provide the key for next-generation personalized decision making.

\printbibliography

@article{bareinboim2014,
    author = "Bareinboim, Elias and Tian, Jin and Pearl, Judea",
    title = "Recovering from Selection Bias in Causal and Statistical Inference",
    journal = "Proceedings of the Twenty-eighth AAAI Conference on Artificial Intelligence",
    address = "Palo Alto, CA",
    publisher = "AAAI Press",
    year = 2014,
    url = "http://ftp.cs.ucla.edu/pub/stat_ser/r425.pdf",
    editor = "Carla E. Brodley and Peter Stone",
    pages = "2410--2416"
}

@article{tian2000,
    title={Probabilities of causation: Bounds and identification},
    author={Tian, Jin and Pearl, Judea},
    journal={Annals of Mathematics and Artificial Intelligence},
    volume={28},
    number={1-4},
    pages={287--313},
    year={2000},
    publisher={Springer},
    url="http://ftp.cs.ucla.edu/pub/stat_ser/r271-A.pdf"
}

@article{pearl2010,
    title={On the Consistency Rule in Causal Inference: An Axiom, Definition, Assumption, or a Theorem?},
    author={Pearl, Judea},
    journal={Epidemiology},
    volume={21},
    number={6},
    pages={872--875},
    year={2010},
    url="https://ftp.cs.ucla.edu/pub/stat_ser/r358-reprint.pdf"
}

@article{pearl2015,
    title={Causes of Effects and Effects of Causes},
    author={Pearl, Judea},
    journal={Journal of Sociological Methods and Research},
    volume={44},
    number={1},
    pages={149--164},
    year={2015},
    url="http://ftp.cs.ucla.edu/pub/stat_ser/r431-reprint.pdf"
}

@article{pearl2014simpson,
    title={Understanding Simpson's Paradox},
    author={Pearl, Judea},
    journal={The American Statistician},
    volume={68},
    number={1},
    pages={8--13},
    year={2014},
    url="http://ftp.cs.ucla.edu/pub/stat_ser/r414-reprint.pdf"
}

@article{rubin1974,
    title={Estimating causal effects of treatments in randomized and nonrandomized studies},
    author={Rubin, Donald B.},
    journal={Journal of Educational Psychology},
    volume={66},
    number={5},
    pages={688--701},
    year={1974},
    doi={10.1037/h0037350}
}

@book{pearl2009causality,
  title={Causality},
  author={Pearl, Judea},
  year={2009},
  publisher={Cambridge University Press},
  edition={{Second}}
}

@misc{mueller2020, title={{Which Patients are in Greater Need}: A counterfactual analysis with reflections on {COVID-19}}, note={\url{https://ucla.in/39Ey8sU+}}, journal={Causal Analysis in Theory and Practice}, author={Mueller, Scott and Pearl, Judea}, year={2020}, month={4}}

@inproceedings{li2019,
  title={Unit selection based on counterfactual logic},
  author={Li, Ang and Pearl, Judea},
  booktitle={Proceedings of the 28th International Joint Conference on Artificial Intelligence},
  pages={1793--1799},
  year={2019},
  organization={AAAI Press}
}

@article{mueller2021,
    title={Causes of Effects: Learning individual responses from population data},
    author={Mueller, Scott and Li, Ang and Pearl, Judea},
    year={2021},
    url="http://ftp.cs.ucla.edu/pub/stat_ser/r505.pdf"
}

@article{dawid2022effects,
      title={Effects of Causes and Causes of Effects}, 
      author={A. Philip Dawid and Monica Musio},
      journal={Annual Review of Statistics and its Application},
      year={2022},
      eprint={2104.00119},
      archivePrefix={arXiv},
      primaryClass={math.ST},
      url="https://arxiv.org/pdf/2104.00119.pdf"
}

@book{porta2016,
      author = "Miquel Porta",
      title = "Number Needed To Treat (NNT)",
      year = "2016",
      publisher = "Oxford University Press",
      isbn = "9780199390069",
      doi = "10.1093/acref/9780199976720.013.1327",
      url = "https://www.oxfordreference.com/view/10.1093/acref/9780199976720.001.0001/acref-9780199976720-e-1327"
}

@article{vancak2020,
author = {V Vancak and Y Goldberg and SZ Levine},
title ={Systematic analysis of the number needed to treat},
journal = {Statistical Methods in Medical Research},
volume = {29},
number = {9},
pages = {2393-2410},
year = {2020},
doi = {10.1177/0962280219890635},
    note ={PMID: 31906795},
URL = { 
        https://doi.org/10.1177/0962280219890635
},
eprint = { 
        https://doi.org/10.1177/0962280219890635
}
}

@article{pearl1999probabilities,
    title={Probabilities of Causation: Three Counterfactual Interpretations and their identification},
    author={Pearl, Judea},
    journal={Synthese},
    volume={121},
    pages={93--149},
    year={1999},
    url="https://ftp.cs.ucla.edu/pub/stat_ser/r260-reprint.pdf"
}

@article {stovitz2013,
	author = {Stovitz, Steven D and Shrier, Ian},
	title = {Medical decision making and the importance of baseline risk},
	volume = {63},
	number = {616},
	pages = {e795--e797},
	year = {2013},
	doi = {10.3399/bjgp13X674585},
	publisher = {Royal College of General Practitioners},
	issn = {0960-1643},
	URL = {https://bjgp.org/content/63/616/e795},
	eprint = {https://bjgp.org/content/63/616/e795.full.pdf},
	journal = {British Journal of General Practice}
}

\end{document}